\documentclass{article}

\newcommand{\etc}{\textit{etc.\@}}

\usepackage[numbers]{natbib}
\usepackage[final]{neurips_2022}

\usepackage[utf8]{inputenc} 
\usepackage[T1]{fontenc}    
\usepackage{hyperref}       
\usepackage{url}            
\usepackage{booktabs}       
\usepackage{amsfonts}       
\usepackage{nicefrac}       
\usepackage{microtype}      
\usepackage{xcolor}         
\usepackage{graphicx}

\title{Dwelling Type Classification for Disaster Risk Assessment Using Satellite Imagery}

\author{%
Md Nasir$^{1}$ \quad Tina Sederholm$^{1}$ \quad Anshu Sharma$^2$ \quad Sundeep Reddy Mallu$^3$ \\
\quad \textbf{Sumedh Ranjan Ghatage}$^3$ \quad \textbf{Rahul Dodhia}$^1$ \quad \textbf{Juan Lavista Ferres}$^1$ \\\
$^1$Microsoft AI for Good Lab \quad $^2$SEEDS \quad $^3$Gramener\\
\texttt{\{mdnasir,tinase,radodhia,jlavista\}@microsoft.com}\\
\texttt{\{anshu\}@seedsindia.org}\\
\texttt{\{sundeep.mallu,sumedh.ghatage\}@gramener.com}
}

\begin{document}

\maketitle

\begin{abstract}
Vulnerability and risk assessment of neighborhoods is essential for effective disaster preparedness.
Existing traditional systems, due to dependency on time-consuming and cost-intensive field surveying, do not provide a scalable way to decipher warnings and assess the precise extent of the risk at a hyper-local level.
In this work, machine learning was used to automate the process of identifying dwellings and their type to build a potentially more effective disaster vulnerability assessment system. First, satellite imageries of low-income settlements and vulnerable areas in India were used to identify 7 different dwelling types. Specifically, we formulated the dwelling type classification as a semantic segmentation task and trained a U-net based neural network model, namely TernausNet, with the data we collected. Then a risk score assessment model was employed, using the determined dwelling type along with an inundation model of the regions. The entire pipeline was deployed to multiple locations prior to natural hazards in India in 2020. Post hoc ground-truth data from those regions was collected to validate the efficacy of this model which showed promising performance. This work can aid disaster response organizations and communities at risk by providing household-level risk information that can inform preemptive actions.
\end{abstract}

\section{Introduction}

In the last two decades, more than 4 billion people worldwide have been affected by disasters. Disasters also affected 1.23 million people and incurred economic losses of 2.97 trillion USD~\cite{1un2020human}.
India is one of the countries at high risk and was ranked the 7th most affected by weather related disasters in 2021 according to the Global Climate Risk Index~\cite{2eckstein2021global}. India loses a large amount of its housing stock to disasters every year, rendering millions of people homeless.
Most of these losses are private and uninsured and are borne by the economically underprivileged communities.
A large part of these losses can be avoided with timely and accurate warnings, and clear advice on actions to be taken to reduce immediate impact during a disaster and building capacity for long term risk reduction. 
The current disaster vulnerability assessments and hazard warnings for communities offer predictions with high variability, and hence low reliability, because they do not consider individual, community, and regional attributes in their forecasts.
In this work, a system to assess risks of naturally triggered hazards was created to improve the granularity of disaster forecasts and advisories.

First we built a machine learning model to detect dwellings and classify them based on rooftop material from satellite imagery.
Valuable information can be inferred by understanding the material used on the roof, including how sound the building structure would be in the case of a naturally triggered hazard such as a flood, cyclone, or earthquake.
Labeled data of two low-income settlements from the Indian cities of Mumbai and Puri were used in training a deep neural network model for identifying buildings and classifying them based on their roof-types.
The problem was formulated as a multiclass semantic segmentation task and aimed to solve by exploring different neural network architectures.
Next, a statistical modeling approach was used for risk scoring to assess the extent of vulnerability.
The on-the-ground knowledge of the \textit{SEEDS} team from working on post-disaster recovery events guided the categorization or risk based on attributes of individual buildings that can be interpreted from rooftop signatures picked through remote sensing using satellite imagery or drone mapping.
This in turn aids vulnerable communities to be better prepared and develop custom response plans.

\section{Application context}

The dwelling type classification serves as the backbone of the risk assessment pipeline we designed and employed.
A roof made of plastic sheet or thatch would have a very high probability of having walls made of bamboo mat or mud with shallow foundations, thus making the entire structure weak; a metal sheet or tiled roof would be interpreted as one having masonry walls with weak mud mortar and inadequate foundations, while a reinforced concrete roof would be assumed to have cement mortar masonry walls with stronger foundations.
Other contributing factors to the damage from disasters also include proximity to water bodies, roads, building footprints, elevation, landslide risks, impervious surfaces, and presence of vegetation cover.

Using our risk assessment model, the risk scores (categorized from 1 to 5) assigned to individual dwelling types, and their concentrations in dwelling clusters, communities can now be provided with accurate warnings at the time of a cyclone that has been forecast by the meteorological department.
While meteorological warnings include cyclonic wind speeds, rainfall and storm surges to be anticipated, the model adds precise information of what this would mean in terms of impact at the site of a specific dwelling, including what damage can be anticipated considering the dwelling type.
With an understanding of their risks, families and communities can take preemptive actions such as evacuation to safer locations and buildings when a disaster is expected in the near future; and strengthening of buildings, raising of water and sanitation structures and formulating disaster preparedness plans to reduce long term risk.  

While models were trained on the data from two cities in India, it was successfully deployed by our nonprofit partner \textit{SEEDS} to two other regions of Tamil Nadu and Kerala, which were impacted by tropical cyclones Nivar and Burevi that made landfall in Southern India in November and December 2020 respectively.  

\section{Dwelling type classification}

The dwelling type classification was approached as a semantic segmentation problem. The goal was to extract structural information of different dwellings in the form of identified polygons.

\subsection{Relevant literature and challenges}

In recent years, building footprint extraction from satellite imagery using machine learning techniques has seen a lot of interests and advances, especially following the SpaceNet challenges~\cite{3van2018spacenet}. For example, \citet{4chartock2017extraction} used a fully convolutional neural network (FCNN) with 11-band image data to extract building footprints. Another CNN-based approach was used for identifying dwellings in refugee camps~\cite{5ghorbanzadeh2018dwelling}. \citet{6zhao2018building} used instance segmentation approach for building detection with a Mask R-CNN.

On the other hand, automatic classification of the dwellings based on their roof material or slope has been rarely studied. A relevant work in this domain used 3-dimensional LiDAR data, which has a lot more information and granularity than satellite imageries, to identify different categories of buildings such as apartment buildings, low-rise buildings, and villas~\cite{7prathap2018deep}. \citet{8lu2014building} used a random forest-based model with LiDAR data to classify buildings. Another work used LiDAR data for identify the shape of the roofs with the application of solar resource assessment~\cite{9huang2017novel}.

This work is the first attempt towards any dwelling type classification from satellite imagery to the best of our knowledge. One of the limitations was the availability of only RGB bands of the satellite imagery. Another major challenge of the work is that it focuses on dense urban landscapes of Indian cities, where the size and shape of the buildings can have significant variation even within a given class.

\subsection{Data collection}

Our work required high quality and reliable satellite imagery with different dwellings identified and labelled based on their roof-type. Free access satellite imagery for Mumbai and Puri region were dated and did not reflect new construction in these cities. To tackle this, recent proprietary satellite imagery was sourced from Maxar Technologies. The imagery obtained for Mumbai and Puri were less than 3 months old with a resolution of 50 centimeter. The resolution of data processed and held locally, as well as the output shared for purpose of warnings was maintained in compliance with local regulations.
Based on the survey of houses in both these regions and the prominent construction methods in use, 12 types of roofs were identified. This was eventually brought down to 7 different types that captured the risk categories which could lead to a simple and clear set of instructions for action. The houses were reviewed manually, using QGIS software, and each house roof type was labelled. A total of 14,000 houses were identified and labeled. To account for inability to identify the right house\/roof type, samples for \textit{no clear roof~(NCR)} and \textit{no clear structure~(NCS)} examples were also included. \textit{No clear roof} refers to the dwellings that do not have a clear view of the roof, but the structure is present; \textit{no clear structure} includes damaged house and houses that cannot be deciphered.  The labeled data served as the base for training our model.

\subsection{Overview of the dataset}

\begin{figure}
  \centering
    \vspace{-1cm}
\includegraphics[height=8cm]{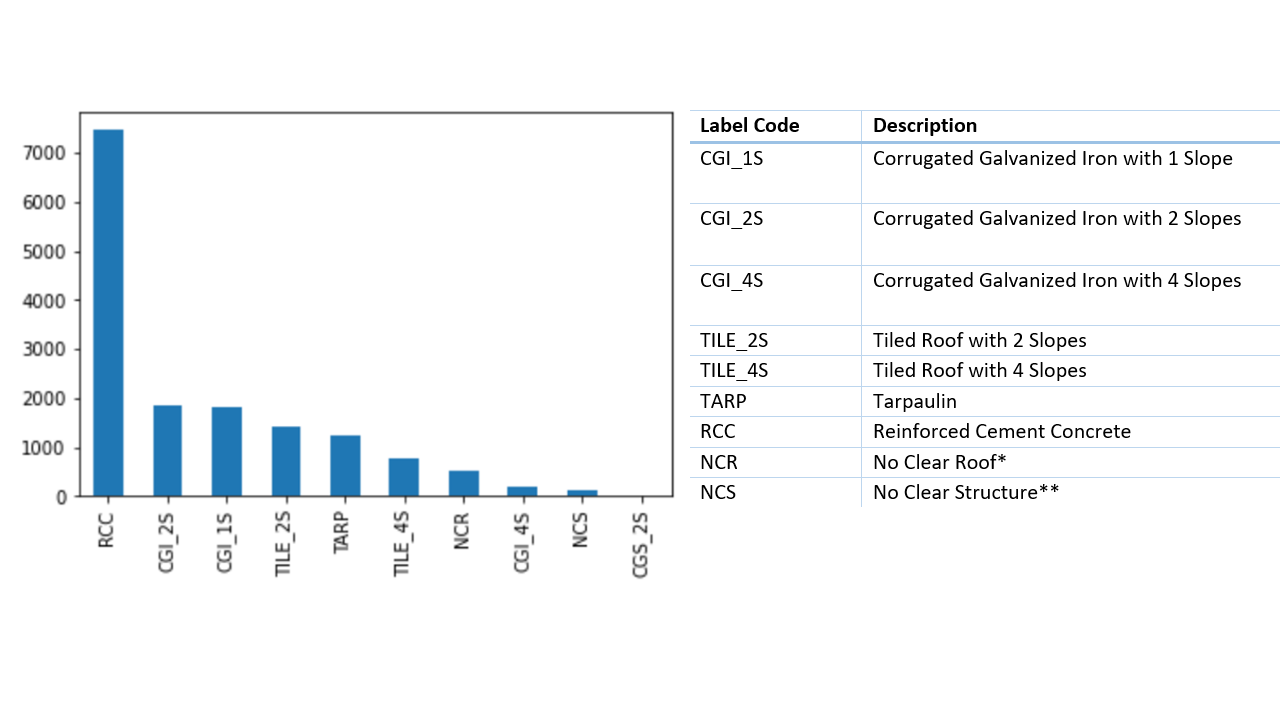}
  \vspace{-2cm}
  \caption{Distribution and description of different dwelling types in the dataset.}
  \label{fig:data}
  \vspace{-0.5cm}
\end{figure}

The dwelling type dataset consists of imagery from 8 areas of interest (AOIs): 4 each from Puri and Mumbai. The class labels for dwelling types and their distribution are shown in Figure \ref{fig:data}.

In the scope of our work, \textit{NCR} and \textit{NCS} categories were excluded since they do not provide any useful information. These categories are defined mostly because of lack of clarity in the satellite imagery. Both \textit{NCR} and \textit{NCS} are ground truthing features and are ignored for modelling as part of the design. While the total number of identified structures were 14746,  the final training dataset had 14109 dwellings after excluding \textit{NCR} and \textit{NCS}.  As evident in the figure, the dataset is highly skewed with a high presence of \textit{RCC} dwelling 
(7,462 counts), while \textit{CGS\_2S} having only 2 instances. This skewness adds to the challenge of imbalance in terms of area (translated as the number of pixels) of dwellings and the background (street, trees, fields, \etc).

\subsection{Preprocessing}
Each of the GeoTIFF images collected from all 8 AOIs were split into smaller tiles of 512x512 pixels, each covering a 256x256 meters on the ground. Each of these image files was accompanied by a GeoJSON label file containing outlines of the dwellings and the label representing their type. The data had 250 such tiles. 36 tiles which did not have any house were discarded, ending up with a total of 214 tiles. The data was randomly split into 70\% training, 15\% validation and 15\% test partitions. The 3-band RGB images created an input size of 512 x 512 x 3.

We experimented with a few data augmentation methods, such as horizontal flipping (reflection) and 90 degrees rotation, random cropping. However, none of the augmentation methods helped improve the performance and hence were not used in the final system. A possible explanation of this behavior could be that the size and orientation of the dwellings of the dataset are diverse enough not to be affected by the augmentation methods.

\subsection{Semantic segmentation Model}

The next step was to train the model in a way that can not only identify pixels that represent dwellings, but also recognize the type of the dwelling (one of the seven categories). The approach to address this problem was to formulate it as multiclass semantic segmentation. More specifically, a neural-network-based model was used to classify each pixel into one of the 8 possible classes: 7 dwelling types and the background class.

\begin{figure}
  \centering
    \vspace{-1cm}
\includegraphics[height=8cm]{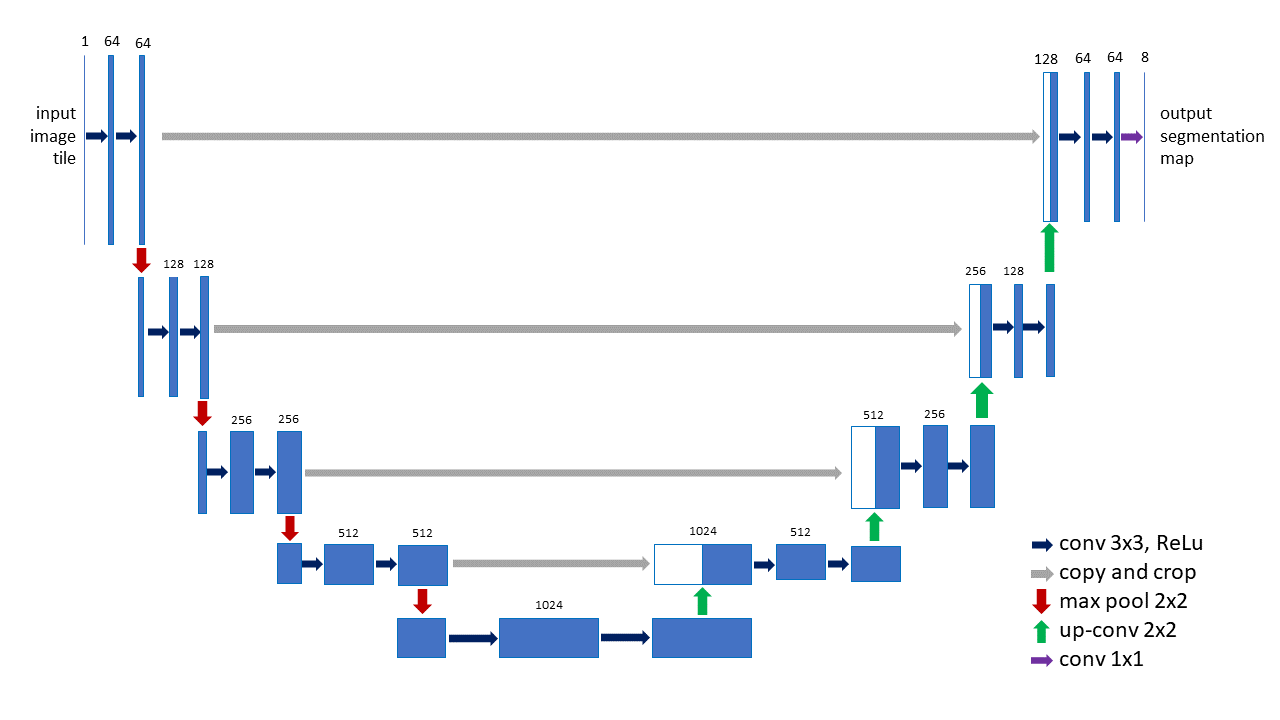}
  \vspace{-1cm}
  \caption{TernausNet (U-Net with VGG11) architecture.}
  \label{fig:arch}
\end{figure}

Motivated by the overwhelming success of U-net~\cite{10ronneberger2015u} based models in semantic segmentation tasks, a recent architecture called TernausNet was adopted~\cite{11iglovikov2018ternausnet}.
The main difference between a standard U-Net and TernausNet is that the latter uses a VGG11 network as the encoder with its weights pre-trained on a larger dataset, ImageNet. The pre-training of the encoder makes the segmentation easier when there is a small dataset to train on, such as the current work.
The original TernausNet was proposed for a binary segmentation task with the goal identifying pixels as \textit{building} or \textit{not building}.
In this work, this approach was extended to the multiclass segmentation by generating an 8-channel output segmentation map, each representing one output class. The architecture is shown in Figure \ref{fig:arch}. Once the segmentation maps are obtained, each pixel is assigned to the class generating maximum value in the corresponding output channel.

\subsection{Training}
To train this network, a multiclass (categorical) cross entropy loss function with one-hot encoded label vector for 8 categories was used. Some other loss functions (such as, multiclass hinge loss, multiclass soft margin loss, and focal loss) and their composite versions were experimented upon; none of which improved the performance. Based on performance on the validation set, we chose to use Adam optimizer with learning rate=0.0001. All experiments were performed using Solaris~\footnote{\url{GitHub - CosmiQ/solaris: CosmiQ Works Geospatial Machine Learning Analysis Toolkit}} toolkit with PyTorch as deep learning framework. The models were trained and tested using 2 Tesla K80 GPUs on an Azure DSVM.

\subsection{Experimental results}
To evaluate the model performance the weighted accuracy based on the class frequency of the pixels has been reported. In addition, we also report the weighted Intersection over Union (IoU) or Jaccard Index. The results of this multiclass semantic segmentation task are shown in Table \ref{tab:results}. As baseline models to compare with the model, we use a fully connected convolutional neural network (FCNN)~\cite{4chartock2017extraction}, a U-Net~\cite{10ronneberger2015u} without any pretrained encoder, and finally a U-Net with a VGG16 encoder pretrained on ImageNet.
In terms of both metrics, the TernausNet model performs the best.

\begin{table}
  \caption{Multiclass semantic segmentation results}
  \label{tab:results}
  \centering
  \begin{tabular}{lll}
    \toprule
    Model     & Weighted Accuracy (\%)     & Weighted IoU  \\
    \midrule
    FCNN                      & $81.92$  & $0.754$     \\
   U-Net (no pretraining)     & $82.30$   & $0.772$       \\
   TernausNet (VGG11 + UNet)  & $88.65$   & $0.809$  \\
    VGG16 + UNet              & $86.98$   & $0.792$  \\
    \bottomrule
  \end{tabular}
\end{table}


\section{Post-processing}
Next, as the post-processing step for the segmentation system, the dwelling footprints obtained from the output maps were polygonized. More specifically, the Douglas-Peucker algorithm~\cite{13douglas1973algorithms, 14ramer1972iterative} while preserving the topology of the shapes was used. This step is necessary not just for smoothing of the dwelling, but also for ensuring a low-resource representation of the dwellings as the input to the risk scoring component, for memory and computational constraints.

\subsection{Impact}
The immediate output of the project is a dramatically improved resolution of flood early warnings. From a set of vague warnings that provide rainfall information for a few square kilometers, the impact of which is hard to understand for communities, more precise information about the risks for a neighborhood can now be obtained. This risk information can be further synthesized to develop advisories and warnings for the vulnerable communities. This contextual information at a much finer resolution enables communities to take preemptive actions to prevent losses. In 2021, the model was run for cyclone Yaas and Tauktae and also for floods in Mumbai.

\section{Conclusions and next steps}
The model is focused on urban risk, as cities are the focus of growing population densities and more complex risks. The project is currently continuing in its second phase, with focus on improving the pipeline and extending its application to multiple hazards besides floods. In addition to improvement of accuracy of the results through more advance modeling, we also aim to improve the reliability of ground truth labels by the use of drone mapping during the annotation process, which provides higher resolution than satellite images.  Our ongoing extension to other hazards primarily includes heatwaves and earthquakes. It may be noted that while dwelling identification modelling approach remains the same with changing hazards, the risk scoring approach changes drastically.  For example, a plastic sheet or tarpaulin roof house may be very weak and vulnerable for floods but will have lower risk of collapse and casualties in an earthquake scenario.  Similarly, a thatch roof (and mud wall) house might have a higher vulnerability in floods but be less vulnerable during an earthquake.  Another improvement to be carried out is the integration of differing categories of hazards and degrees of damage to houses.  
By incorporating attributes unique to each community, such as building materials and topography, the project aims to create a more accurate community-specific model to predict vulnerability to natural hazards, allowing communities to be better prepared in advance and develop custom response plans.  Our disaster response partner, \textit{SEEDS}, is also working towards policy engagements to make the model available to local, state, and national government agencies for mainstreaming and scaling up, striving to see it applied nationally to make the lives of millions safer.

\begin{ack}
This work was funded by AI for Humanitarian Grant under AI for Good initiative of Microsoft. The authors would like to thank Shubham Mishra and Sreya Ajay from SEEDS for their contributions.
\end{ack}

\bibliographystyle{abbrvnat}
\bibliography{references}

\appendix

\newpage

\section{Appendix}

\begin{figure}
  \centering
\includegraphics[height=20cm]{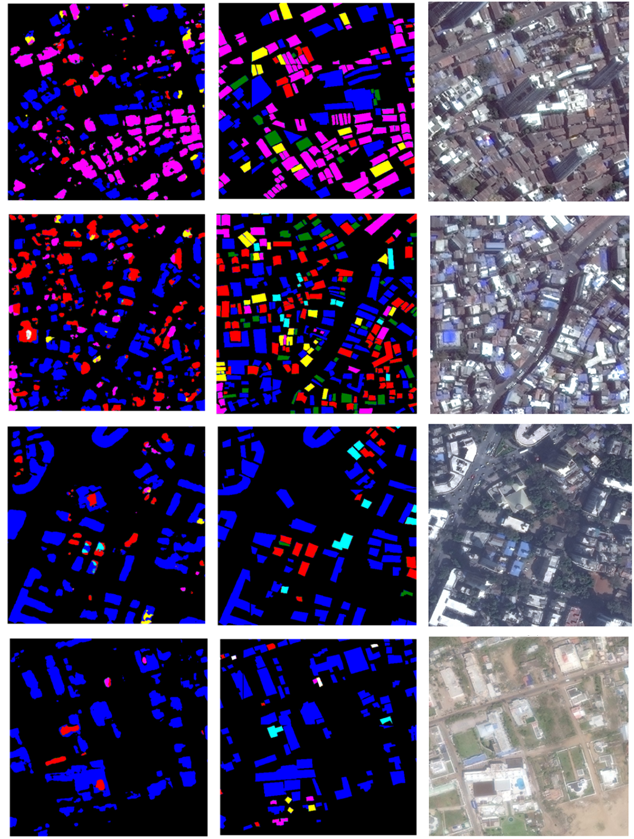}
  
  \caption{Some examples of the TernausNet model prediction of dwelling type \\
(Left column): predicted label from the model output, (Middle): ground truth, (Right): input imagery.  Different colors in the first columns indicate different dwelling types, except black which represents the background (no dwelling)}
  \label{fig:sample}
\end{figure}


\begin{table}[h]
  \caption{Binary segmentation (dwelling vs. no dwelling classification) results}
  \label{tab:resultsbin}
  \centering
  \begin{tabular}{llll}
    \toprule
    Model     & Accuracy (\%)     & Precision & Recall  \\
    \midrule
    FCNN                      & $89.49$  & $0.872$   & $0.826$     \\
   U-Net (no pretraining)     & $86.97$  & $0.782$   & $0.899$  \\
   TernausNet (VGG11 + UNet)  & $90.96$  & $0.893$   & $0.800$  \\
    VGG16 + UNet              & $88.15$  & $0.854$   & $0.831$  \\
    \bottomrule
  \end{tabular}
\end{table}

\begin{figure}
\includegraphics[height=4cm]{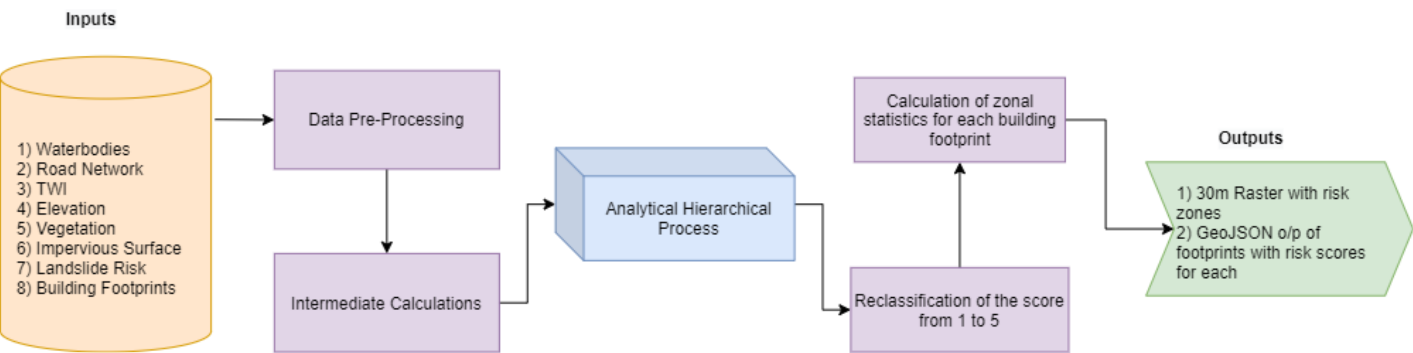}
  
  \caption{Risk score modelling pipeline overview}
  \label{fig:pipe}
\end{figure}

\end{document}